\pdfoutput=1

\documentclass[letterpaper, 10pt, conference]{ieeeconf}      
\IEEEoverridecommandlockouts                              
\usepackage{multicol}
\usepackage{multirow}
\usepackage{algpseudocode}
\usepackage{color}
\usepackage{xcolor}
\usepackage{graphicx}
\usepackage{bm}
\usepackage{amsmath}
\usepackage{amssymb}
\usepackage{tabularx}
\usepackage{subcaption}
\usepackage{url}
\usepackage{cite}
\usepackage{prettyref}
\usepackage{booktabs}
\usepackage{siunitx}
\usepackage{comment}

\usepackage{hyperref}
\usepackage[all]{hypcap}

\usepackage[ruled]{algorithm2e} % For algorithms

\SetAlFnt{\small}
\SetAlCapFnt{\small}
\SetAlCapNameFnt{\small}
\SetAlCapHSkip{0pt}
\newrefformat{fig}{Figure~\ref{#1}}
\newrefformat{sec}{Section~\ref{#1}}
\newrefformat{tab}{Table~\ref{#1}}
\newrefformat{alg}{Algorithm~\ref{#1}}
\newrefformat{eqn}{Equation~\ref{#1}}

\hypersetup{
    colorlinks,
    linkcolor={blue!80!black},
    citecolor={blue!80!black},
    urlcolor={blue!100!black}
}

\makeatletter
\newcommand{\printfnsymbol}[1]{%
  \textsuperscript{\@fnsymbol{#1}}%
}
\makeatother

\title{Collision-Aware Fast Simulation for Soft Robots by Optimization-Based Geometric Computing}
\author{Guoxin Fang$^*$, Yingjun Tian$^*$, Andrew Weightman, and Charlie C.L. Wang$^\dag$
\thanks{All authors are with the Department of Mechanical, Aerospace and Civil Engineering, The University of Manchester, United Kingdoms}
\thanks{Guoxin Fang is also a PhD student at the Faculty of Industrial Design Engineering, Delft University of Technology, The Netherlands.}
\thanks{$^*$Denotes equal contribution.}
\thanks{$^\dag$Corresponding author: {\tt\small changling.wang@manchester.ac.uk}}
}

\begin{document}
\maketitle
%\thispagestyle{empty}
%\pagestyle{empty}

%%%%%%%%%%%%%%%%%%%%%%%%%%%%%%%%%%%%%%%%%%%%%%%%%%%%%%%%%%%%%%%%%%%%%%%%%%%%%%%%
\begin{abstract}
Soft robots can safely interact with environments because of their mechanical compliance. Self-collision is also employed in the modern design of soft robots to enhance their performance during different tasks. However, developing an efficient and reliable simulator that can handle the collision response well, is still a challenging task in the research of soft robotics. This paper presents a collision-aware simulator based on geometric optimization, in which we develop a highly efficient and realistic collision checking / response model incorporating a hyperelastic material property. Both actuated deformation and collision response for soft robots are formulated as geometry-based objectives. The collision-free body of a soft robot can be obtained by minimizing the geometry-based objective function. Unlike the FEA-based physical simulation, the proposed pipeline performs a much lower computational cost. Moreover, adaptive remeshing is applied to achieve the improvement of the convergence when dealing with soft robots that have large volume variations. Experimental tests are conducted on different soft robots to verify the performance of our approach.
\end{abstract}

% Keywords appear just beneath the abstract. Use only for final RAL version.  
% \begin{IEEEkeywords}
% collision-aware simulation; geometry computing; modeling, control, and learning for soft robots
% \end{IEEEkeywords}

\section{Introduction}
\label{sec:intro}

\begin{figure}[t]
\centering
\includegraphics[width=\linewidth]{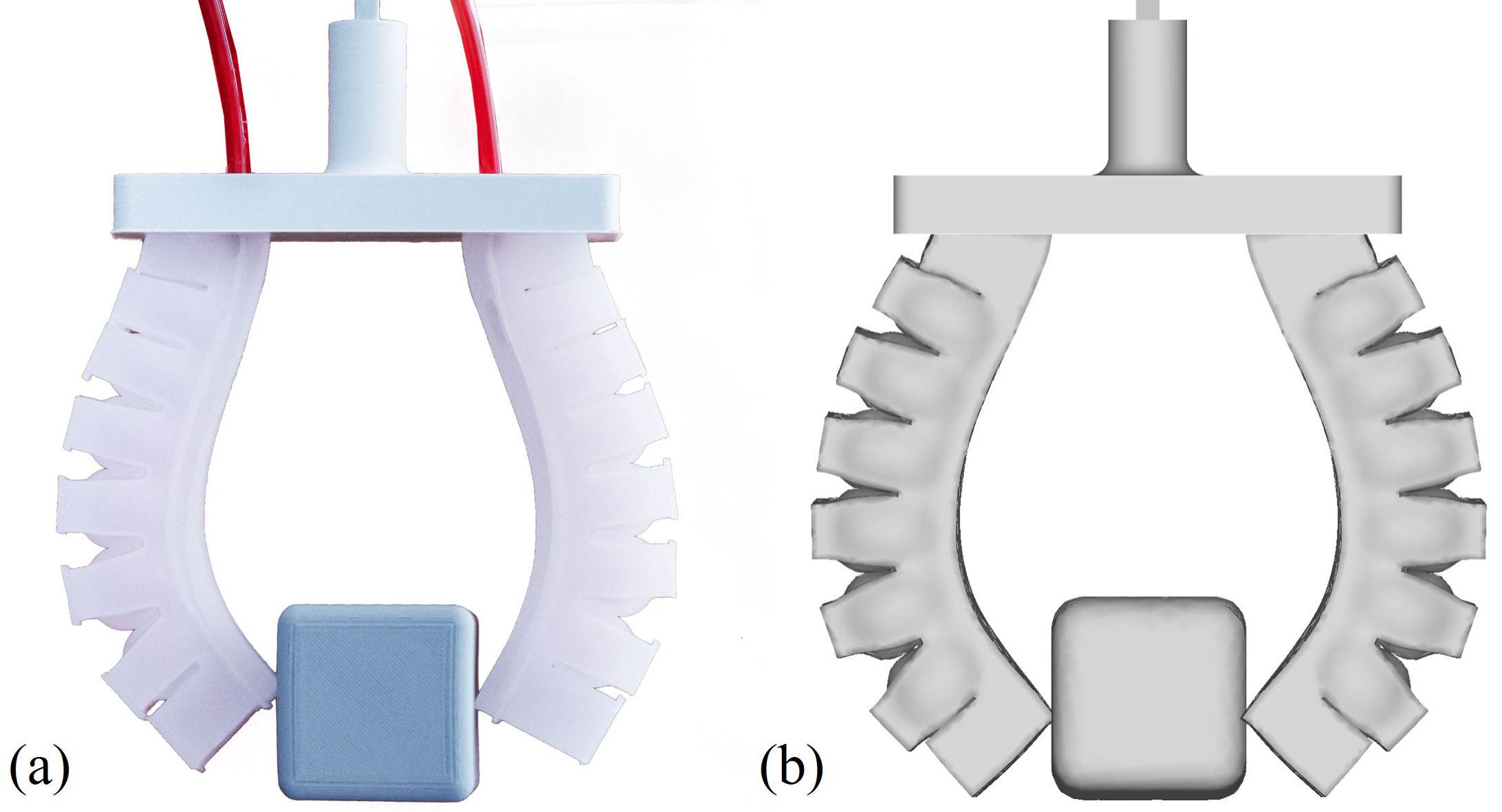}
\caption{Pneumatically actuated soft gripper that can effectively grasp objects by large inflation of chambers and the self-collision between neighboring chambers: (a) the physical result on a soft gripper made by silicone casting and (b) our simulation result that can well predict the shape of the soft gripper by considering both the robot-robot and the robot-obstacle collisions.
%(a) Pneumatically actuated soft gripper that can apply effective grasping of external object with the help of large chamber expansion and self-collision in 'teeth' structure. (b) Our simulator can well predict the shape of the soft robot by considering both robot-robot and robot-obstacle interactions. 
}\label{fig:teaser}
\end{figure}

% -- brief introduction
Powered by the flexibility of soft materials and novel structure designs, soft robots are able to perform complex deformation in their body shapes and safely interact with other objects~\cite{rus2015design}. Thus, they have many advantages in certain tasks, such as grasping fragile objects and exploring confined environments~\cite{galloway2016soft, connolly2015mechanical}, which are challenging for conventional robots that have rigid bodies. When controlling rigid robots to complete tasks, robot-robot collisions and robot-environment collisions are generally prohibited. A simplified solution is to avoid contact. However, this is not the case for soft robots since the phenomenon of contact caused by collision plays an important role in allowing them to safely interact with environments and their own bodies. One should note that self-collision has been employed in modern design to achieve advanced functionality~\cite{marchese2015recipe, morales2020design}. For example, the soft gripper presented in Fig.~\ref{fig:teaser} has the ability to interact and grasp objects with its soft body. Meanwhile, the self-collision that happens in the gap region between chambers can enhance the stiffness of the grasping and provide faster responses in bending deformation (ref.~\cite{mosadegh2014pneumatic, goury2021real}).

\subsection{Challenges in Collision-Aware Simulation for Soft Robots}
At present, the design process of soft robots heavily relies on the experience of engineers. Building effective computational tools is essential for accelerating the process. Studies have been conducted on using Finite-Element Analysis (FEA) simulation in the loop of design optimization (e.g.,~\cite{moseley2016modeling, chen2019optimal}). Meanwhile, predicting the behavior of soft robot under actuation also plays an important role in building fast model-based controllers~\cite{thuruthel2018model}. 

Efficiently simulating the deformation of soft robots while considering self-collisions and interactions with obstacles is still a very challenging problem. Incorporating the collision response into the simulation of soft robots requires to know their whole body shapes. In this case, an FEA-based simulator is more general and effective than the analytical models (e.g., \cite{zhong2021bending}). Nevertheless, hyperelastic material properties and the highly nonlinear deformation presented on soft robots~\cite{marchese2016design, yang2020twining} (e.g., expanding, twisting, and bending) make obtaining an accurate result from simulations time-consuming. 

The challenges of building collision-aware simulation for soft robot are summarized in the following list.
\begin{enumerate}
\item \textit{Complexity in deformation:} The involvement of highly nonlinear deformation makes FEA-based simulations difficult to converge in numerical iterations~\cite{FANG20_TRO}. For example, the largely inflated chambers for pneumatic-driven soft robots can lead to highly-distorted elements for FEA, which introduces numerical instability. 

\item \textit{Fast collision checking through actuation:} Unlike articulated robots with rigid bodies, the freeform deformable body shape of a soft robot makes collision detection and self-collision detection more difficult to conduct efficiently.

\item \textit{Efficient collision response with hyperelastic material properties:} Once (self-)collisions are detected, a response algorithm needs to modify a soft robot's current shape to eliminate collided regions. Providing effective collision response while being able to mimic the complex physical behaviors of hyperelastic materials is challenging.
\end{enumerate}
In this work, we propose a method of optimization-based geometric computing to tackle the mentioned challenges. 

\subsection{Related Work}
To model the behavior of soft robot with interaction, full FEA-based simulations are most widely used due to their proofed high accuracy~\cite{polygerinos2015modeling, moseley2016modeling}. However, the computational cost is high since a small time step needs to be used to solve the nonlinear geometry change (mainly rotational) with collision response. Reduced FEA-based numerical models that are based on voxel~\cite{hiller2014dynamic} or particle~\cite{hu2019chainqueen} approximation can achieve simulation with real-time speed. Based on off-the-shelf physics engine of rigid-body simulation, Graule \textit{et al.}~\cite{graule2021somo} built a model to predict the behavior of a continuum robot finger when interacting with a complex environment.
The self-collision issue was not considered in these frameworks. A general simulator based on force-equilibrium function was developed by Duriez \textit{et al.}~\cite{duriez2013control}, and it can achieve the speed of real-time simulation with the help of model reduction~\cite{goury2018fast}. This work was recently enhanced by the self-collision response and has improved its accuracy when controlling cable-driven soft robots~\cite{goury2021real}. However, the (self-)collision region was defined during the pre-processing step and was fixed through actuation. Furthermore, their method based on integration along time steps accumulates the error; therefore, the simulation could become non-realistic for pneumatic-driven soft robots with large expanding and complex rotational deformation, %(comparison and discussion presented in Sec.~\ref{sec:result}). 
which has been discussed in \cite{FANG18_ICRA}.

For applications such as cloth simulation and animation, research has been conducted in computer graphics (e.g.,~\cite{dinev2018FEPR,projectiveDynamics,li2020soft}) to tackle the problem of collision detection and response for deformable objects. Methods based on \textit{bounding-volume hierarchies} (BVHs), distance fields, and spatial partitioning were utilized for fast collision detection (see~\cite{Teschner_CGF_review} for a comprehensive review). In this work, BVH-based acceleration is applied to support real-time (self-)collision detection for soft robots. For collision response, the method presented in \cite{volumeContactConstraint} was based on using constrained optimization to minimize the collided region between two objects. 
%However, solving the constrained optimization problem is time-consuming. \charlie{I am not sure about this; suggest to remove this sentence.}\guoxin{I agree, please remove this.}
Penalty force-based method was also invited to accelerate the collision-response process with time integration~\cite{Terzopoulos}, which however suffers from the difficulty of tuning parameters to avoid unrealistic results in simulations. Meanwhile, none of these works incorporates the hyperelastic properties of materials that are commonly used to fabricate soft robots. In this paper, we bring the idea from spring-mass system~\cite{Liu:2013:FSM} and build the `virtual' spring elements for collision response. When the spring energy for these elements is minimized, collision-free is achieved on the bodies of soft robots. 
%
%M. Teschner\cite{deformableObjectDetection} systematically introduced different types of deformable collision detection methods such as bounding-volume hierarchies (BVHs), distance fields, spatial partitioning and more. Especially, BVHs have proven to be among the most efficient data structures for collision detection. In this paper, AABB tree which is one type of the BVHs \cite{AABB} \cite{bspBoundaryExt} has been used to detect the collision information including self collision and collision with environment. In collision response, there are three ways to handle collisions: penalty force\cite{Terzopoulos}\cite{contactMech}, impulse reaction\cite{impulseResponse}\cite{velocityFilter} and constraints\cite{volumeContactConstraint}\cite{InterferenceAware}.
%

% discuss the material usage
%To enable a softness in body , material with high elasticity (e.g., silicon rubber or ..) is used to fabricate soft robot, which allows more than 20x expansion in the chamber when actuated.

\subsection{Our Method}
In this work, we present an optimization-based simulator that can efficiently predict complicated deformation for soft robots and effectively provide realistic collision responses by incorporating hyperelastic material properties. Our method extends the pipeline of geometry computing previously presented in~\cite{FANG18_ICRA,FANG20_TRO}. The elastic energy in soft body is minimized to compute its deformed shape, and the actuation is formulated as constraints in terms of length, area or volume variation applied to some elements. After using axis-aligned bounding boxes (AABB) as BVH to support real-time (self-)collision detection, we introduce a group of 'virtual' spring elements to link collided points with their collision-free corresponding vertices. The hard-to-solve collision response constraint can then be reformulated as a quadratic energy term to be minimized. A local-global iterative solver is employed to find the collision-free shape under certain actuation. The geometry-based  model~\cite{Christ21_JCISE} is adopted in our approach for hyperelastic materials, the parameters of which can be obtained from calibration.  

The technical contributions of our work include the following:
\begin{itemize}
    \item A geometry-based simulator that can effectively predict the shape of %pneumatic-driven 
    soft robots with large and complex deformation by incorporating adaptive remeshing;
    
    \item A fast collision checking method to detect both robot-robot and robot-obstacle interactions through the deformation;
    
    \item A spring-element assisted collision response model that incorporates hyperelastic material properties.
\end{itemize}
The generality and performance of our collision-aware soft robot simulator have been tested and verified on different soft robots fabricated from different materials. In all experimental tests, our simulator can successfully mimic the physical behaviors of soft robots with high accuracy. When compared with FEA-based methods, our simulator demonstrated better performance in its efficiency and numerical convergence capability. The code of our simulator is open-source at \url{https://github.com/YingGwan/collisionAwareSOROSimulator}.
\section{Geometry-based Simulation for Soft Robot with Collisions}
This section first presents how to formulate the problem of predicting the deformed shape of soft robot under certain actuation as a constrained optimization problem that can be solved by geometric computing. The fast collision detection is enabled by BVH based search. The collision response model will be formulated by adding `virtual' spring elements into the constrained optimization framework. %Details are presented below.

\subsection{Geometry-based optimization for modelling soft robots}
\begin{figure}[t]
\centering
\includegraphics[width=\linewidth]{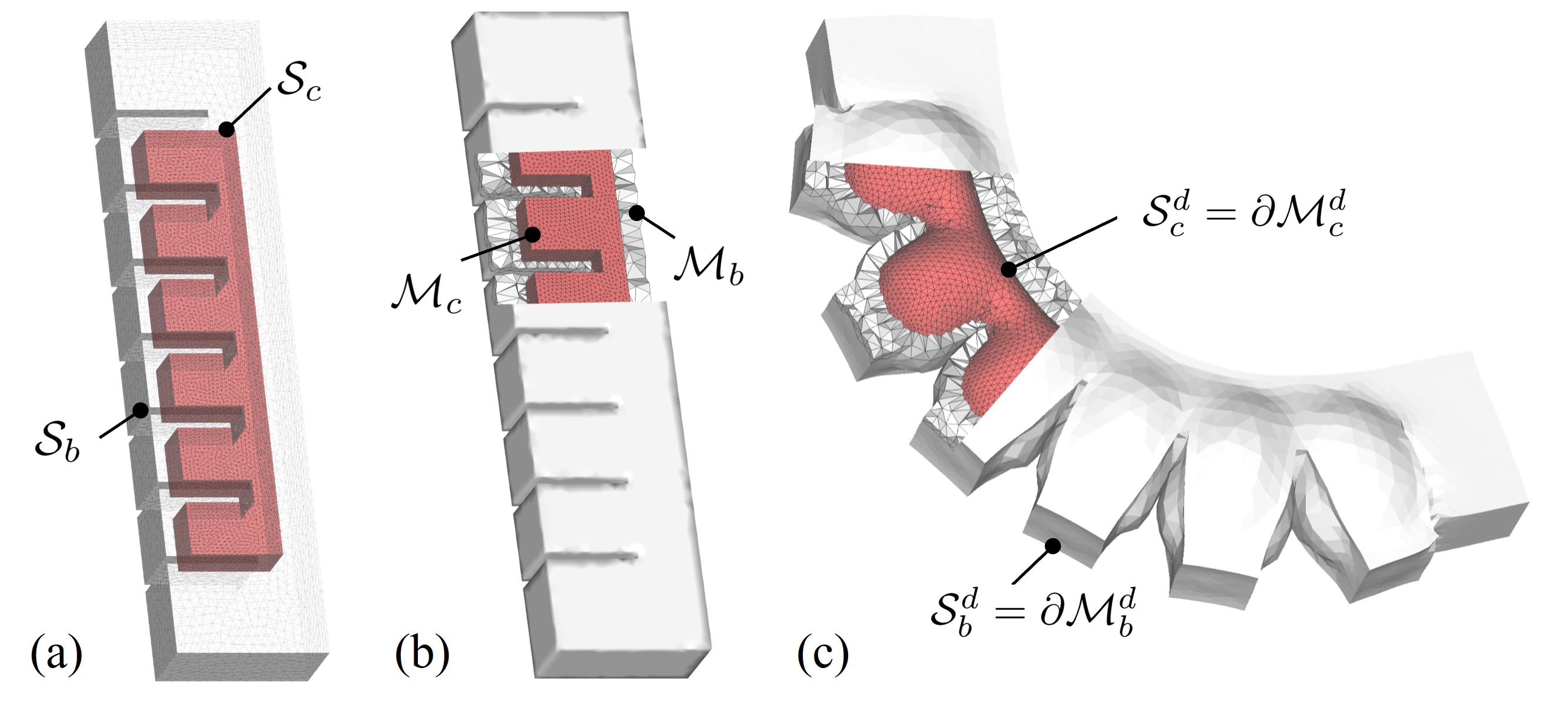}
\caption{(a) The input shape of a soft robot is represented by two surface meshes $\mathcal{S}_c$ (as chamber) and $\mathcal{S}_b$ (as body). (b) Volume tessellation is applied to generate tetrahedral meshes $\mathcal{M}_c$ and $\mathcal{M}_b$. (c) Inflation in the chamber region drives the soft robot being deformed to $\mathcal{M}^d = \{\mathcal{M}^d_c, \mathcal{M}^d_b\}$. Adaptive remeshing is applied to update $\mathcal{M}_c$ during the deformation.}\label{fig:pipeline}
\end{figure}

As illustrated in Fig.\ref{fig:pipeline}, we require an input of two surface meshes $\mathcal{S}_{c}$ and $\mathcal{S}_{b}$ that represent the initial shape of a soft robot in the form of its chamber and body, respectively. Volumetric tessellation (function remarked as $\mathcal{T}(\cdot)$) is applied to generate a tetrahedral mesh that interpolates these two surface meshes. Note that the tetrahedral elements are generated between $\mathcal{S}_{c}$ and $\mathcal{S}_{b}$ and also inside $\mathcal{S}_{c}$. $\mathcal{M}$ is segmented into two topologically connected parts $\mathcal{M}_c$ and $\mathcal{M}_b$. Elements in $\mathcal{M}$ are categorized as \textit{chamber element} ($e\in\mathcal{M}_c$, depicts in red) and \textit{body element} ($e\in\mathcal{M}_b$, depicts in gray). We remark the shape of each element by matrix $\mathbf{V}_{e} = [\mathbf{v}_1, \mathbf{v}_2, \mathbf{v}_3, \mathbf{v}_4] \in \mathbb{R}^{3 \times 4}$, where $\mathbf{v}_i$ gives the position of the i-th vertex.

%Here the function $\mathcal{T}(\cdot)$ can guanatee the protection of the input model, therefore the connection between boudary of  can be build as $\partial\mathcal{M}_c = \mathcal{S}_c$ and $\partial\mathcal{M}_b = \mathcal{S}_b$. Here the the boundary vertex is protected therefore we can ensure the boundary of the tetrahedral mesh $\partial\mathcal{M}_c = \mathcal{S}_c$ and $\partial\mathcal{M}_b = \mathcal{S}_b$.

Without loss of generality, when pneumatic actuation is applied, the deformation of soft robot is driven by the volume expansion in chamber elements here\footnote{For the actuation based on length or area variations, they can be formulated in a similar way. Details can be found in \cite{FANG20_TRO}.}. Meanwhile, the shape of body region will be changed as the elements are topologically connected. Here a target volume expansion ratio $\alpha$ is given as the input actuation parameter. In the geometry-based simulation pipeline~\cite{FANG20_TRO}, shapes of the body elements were computed by minimizing the elastic energy $E_{ela}$. This energy is evaluated by the shape difference between the element's \textit{deformed shape} $\mathbf{V}^d$ and its \textit{target shape} $\mathbf{V}^t$ (details will be presented in Sec.~\ref{subsec:localProjection}). Meanwhile, collisions can occur either between different bodies of the soft robot (i.e., self-collision) or between the soft robot and external obstacles.

The optimization problem that computes deformed shape $\mathcal{M}^d$ is formulated as 
\begin{align}
\arg \min_{\mathcal{M}^d} \quad & E_{ela} = \sum_{e \in \mathcal{M}_b} \mathit{Vol}(e) \|\mathbf{N}\textbf{V}^d - \mathbf{R} (\mathbf{N}\textbf{V}^t) \|^2  \label{eq:eleEnengy} \\
s.t. \quad & \mathit{Vol}(\mathcal{M}_c^d) =  \alpha\sum_{e\in\mathcal{M}_c}\mathit{Vol}(e)  \label{eq:actuConst} \\
& \mathbf{v} \cap e = \emptyset \;  (\forall \mathbf{v} \in \mathcal{S}_c, e\in \mathcal{M})  \label{eq:interaction} \\
&  \mathcal{M} \cap \mathcal{O}_{obs} = \emptyset  \label{eq:selfCollision}
\end{align}
In the definition of elastic energy (Eq.(\ref{eq:eleEnengy})), the matrix $\mathbf{N} = \mathbf{I}_{4\times4} - \frac{1}{4}\mathbf{1}_{4\times4}$ is used to transform the element to the coordinate origin. Rotation matrix $\mathbf{R}$ is computed by \textit{single value decomposition} (SVD) and aligns $\mathbf{V}^d$ and $\mathbf{V}^t$ to the same pose.
The operator $\mathit{Vol}(\cdot)$ in the actuation constraint (Eq.(\ref{eq:actuConst})) computes the volume of given element. 
The self-collision-free requirements are formulated as hard constraints (Eq.(\ref{eq:interaction})) where each vertex located at the boundary of $\mathcal{M}_b$ (remark as $\mathbf{v}\in \partial\mathcal{M}_b = \mathcal{S}_c$) should not go inside any of the other tetrahedral elements on $\mathcal{M}$.
%Collision constrains (\ref{eq:interaction}) ensure that all the vertex located at the boundary of $\mathcal{M}_b$ should not intersect with the soft robot model $\mathcal{M}$ or external obstacle $\mathcal{O}_{obs}$.
Interaction with the obstacles are handled by the last hard constraint (Eq.(\ref{eq:selfCollision})).

To effectively solve this nonlinear system, a projection-based local-global solver is introduced in our previous work~\cite{FANG20_TRO}; however, we need to deal with the newly added collision constraints here (i.e., Eqs.(\ref{eq:interaction}) and (\ref{eq:selfCollision})). The details of this solver incorporating collision response are presented in Sec.~\ref{sec:solver}. We first discuss the method for fast collision detection and our collision response model in the following section.
%We invite the spring element to transfrom the collision constrain as objective function that can be adaptively minimized by a similar projection-based solver (presented in next section).

\begin{figure}[t]
\centering
\includegraphics[width=0.8\linewidth]{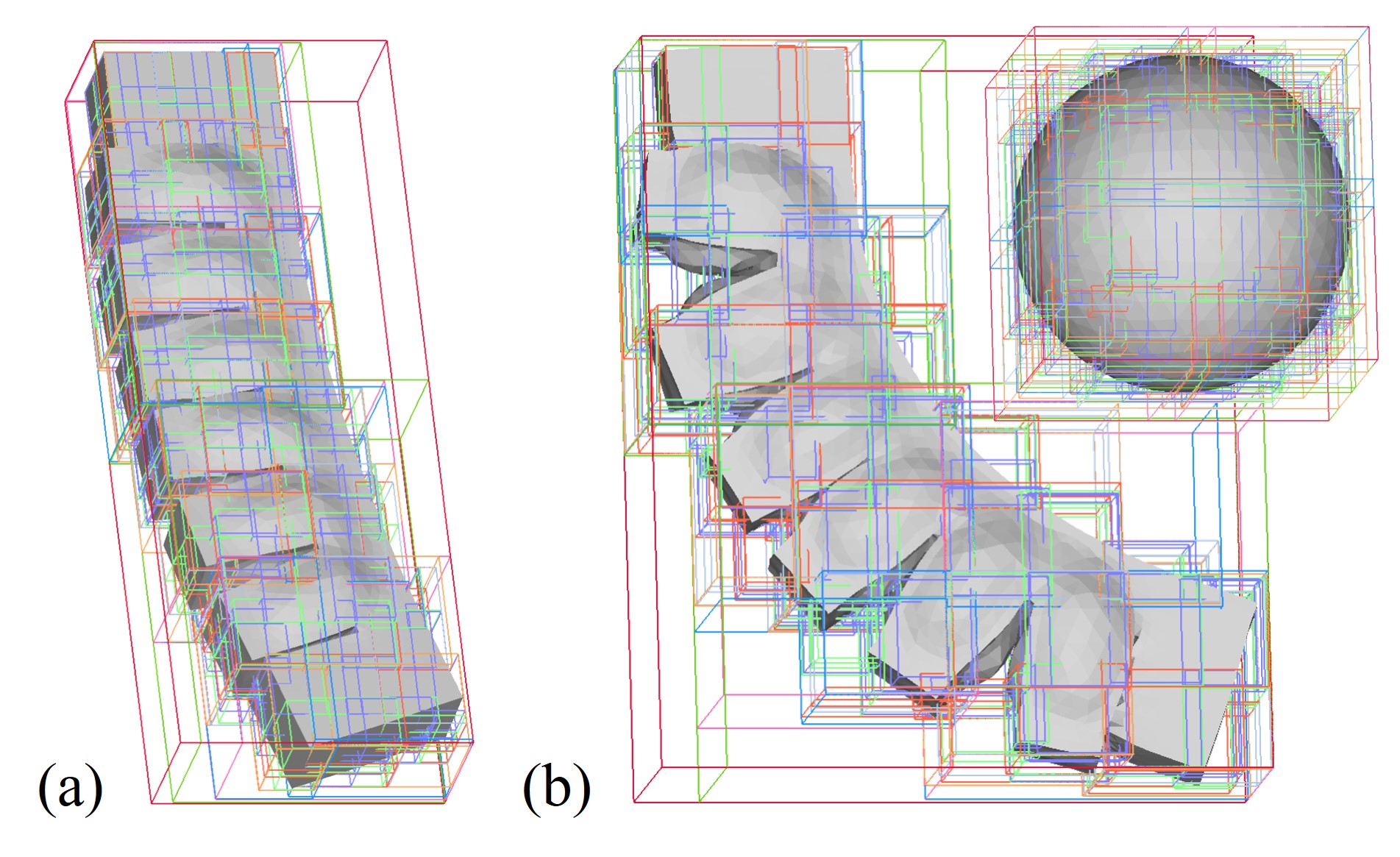}
\caption{(a) AABB trees are built for both soft robot and obstacle for fast collision detection. (b) The structure of BVHs can be re-used when the deformation is relatively small -- i.e., only the dimensions of the bounding-boxes need to be updated.}\label{fig:AABBTree}
\end{figure}

\subsection{Fast (self-)collision detection with BVHs}
For the purpose of collision detection, a brute-force solution is to check all tetrahedra for every vertex on the surface of a soft robot to see if any intersection happens. This method however introduces a high computational cost when dense meshes are involved. An algorithm based on BVHs is utilized to accelerate the search by avoiding unnecessary checks of collision pairs during the computation.

Firstly, the \textit{axis-aligned bounding box} (AABB) trees~\cite{Gino98_AABB} are constructed for both $\mathcal{M}$ and $\mathcal{O}_{obs}$. The bounding-box of a whole model is set as the tree's root, and the nodes in different levels are built to split elements into smaller and smaller bounding boxes until reaching the leaf nodes that only contain one tetrahedron (illustrated in Fig.~\ref{fig:AABBTree}). Both the self-collision and obstacle interaction can be efficiently detected by the traversal on the AABB trees, which has the complexity of $\mathcal{O}(n \log(n))$ with $n$ being the number of elements. 

AABB trees are employed in our framework instead of other BVHs since the structure of the tree can be kept through the deformation while only the geometry of bounding boxes is updated by the new positions of tetrahedra. The tightness of bounding will become loose after presenting large deformation on the soft bodies. The trees can be re-built when necessary. In all our examples presented in this paper, we only rebuild the AABB trees after remeshing. The collision detection can always be efficiently conducted on AABB trees without re-building. Based on our test, (self)-collision detection can be completed at 24fps for a mesh that contains 40k tetrahedra. In each iteration, we stored all collided vertices in the set $\mathcal{V}_{col}$. Statistics regarding the computational time for our collision detection are listed in Sec.~\ref{sec:result}.

\begin{figure}[t]
\centering
\includegraphics[width=1.0\linewidth]{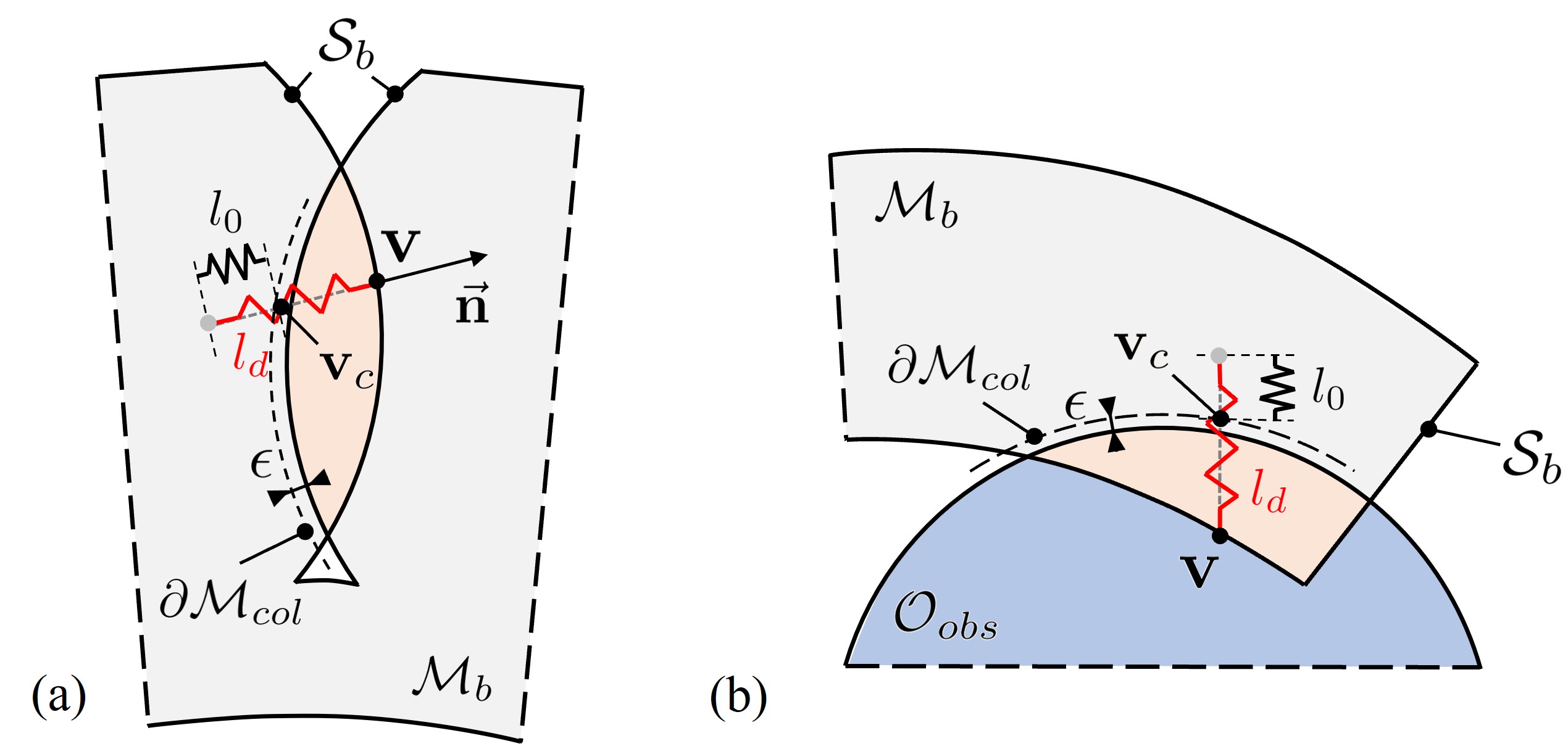}
\caption{Illustration of building `virtual' spring elements for collision response. The starting point of a spring can be conceptually considered as a point through the direction of $\Vec{\mathbf{vv}_c}$ inside the model. (a) Self-collision happens between two chamber regions that both belong to the body mesh $\mathcal{M}_b$. (b) Interaction between $\mathcal{M}_b$ and $\mathcal{O}_{obs}$.}\label{fig:collisionSpring}
\end{figure}
 %The element is elongated from initial length $l_0$ to its deformed status $l_d$ (defined by $\mathbf{v}$ and its correspondence point $\mathbf{v}_c$) due to the collision. 

\subsection{Collision response by spring elements}\label{subsec:collResponse}
After detecting the collided regions, we build virtual springs on every vertex $\mathbf{v} \in \mathcal{V}_{col}$ for the purpose of collision response. As illustrated in Fig.~\ref{fig:collisionSpring}, when a vertex $\mathbf{v}$ goes inside the collided region (depicted in orange), we can consider adding a virtual spring between $\mathbf{v}$ and the predicted position $\mathbf{v}_c$ where it should contact the interface after removing collision. Without loss of generality, we can conceptually assume the spring's length was changed from its resting length $l_0$ to its deformed status $l_d$. To compute the collision-free shape using geometric optimization, the `virtual' spring will drag the collided vertex $\mathbf{v}$ moving out to a collision-free position $\mathbf{v}_c$. This collision response can be formulated as minimizing the spring energy:
\begin{equation}
    E_{col} = k(l_d - l_0)^2 = k \|\mathbf{v}-\mathbf{v}_c \|^2.
\end{equation}
When $E_{col}$ is minimized close to zero, the vertex would no longer collide with the model itself or other obstacles. 

For the case of self-collision as illustrated in Fig.~\ref{fig:collisionSpring}(a), the corresponding position $\mathbf{v}_c$ is computed by searching along the inverse normal direction $-\Vec{n}(\mathbf{v})$ from $\mathbf{v}$ until intersecting with the boundary of collided region (remarked as $\partial \mathcal{M}_{col}$). In the case of robot-obstacle interaction, $\mathbf{v}_c$ is found to be the closest point on $\partial \mathcal{M}_{col}$ (see Fig.~\ref{fig:collisionSpring}(b)). To avoid numerical error, $\partial \mathcal{M}_{coll}$ is computed by giving a small offset $\epsilon$ to the boundaries of robots and obstacles. 
%The closest point can be efficiently computed with the help of the PQP library.
%Meanwhile, $\mathbf{v}_c$ is in general detected at the middle of one triangle face $f_c\in\partial \mathcal{M}_{coll}$, therefore barycentric coordinate is applied to precisely define the position of $\mathbf{v}_c$ to the three vertices of $f_c$. 

During the iteration of deformation, corresponding points need to be updated accordingly. The spring elements also need to be added or removed through the iteration to avoid unrealistic adhesion behavior. In the next section, we will present the details of adaptive collision response and a local-global solver that minimizes $E_{ela} + E_{col}$ together with the actuation constraints (i.e., Eq.(\ref{eq:actuConst})).

\section{Projection-Based Solver with Remeshing}\label{sec:solver}
By incorporating collision constraints as the spring energy to be minimized, the collision-aware simulation for soft robots is reformulated as an optimization problem. A local-global iterative solver is used to solve this system. In the local step, the target shapes for chamber elements, body elements, and spring elements are computed by local projection. These target shapes are set to satisfy the corresponding objectives or constraints. In this local step, collision detection is also applied to update spring elements accordingly. After that, a global blending step is applied to assembly all elements according to their topological connections. Repeatedly applying the local projection and the global blending steps results in the collision-free shape of a soft robot. Meanwhile, progressive remeshing is conducted to ensure a stable and realistic result when simulating soft robots with very large chamber deformation.

\begin{figure}[t]
\centering
\includegraphics[width=0.9\linewidth]{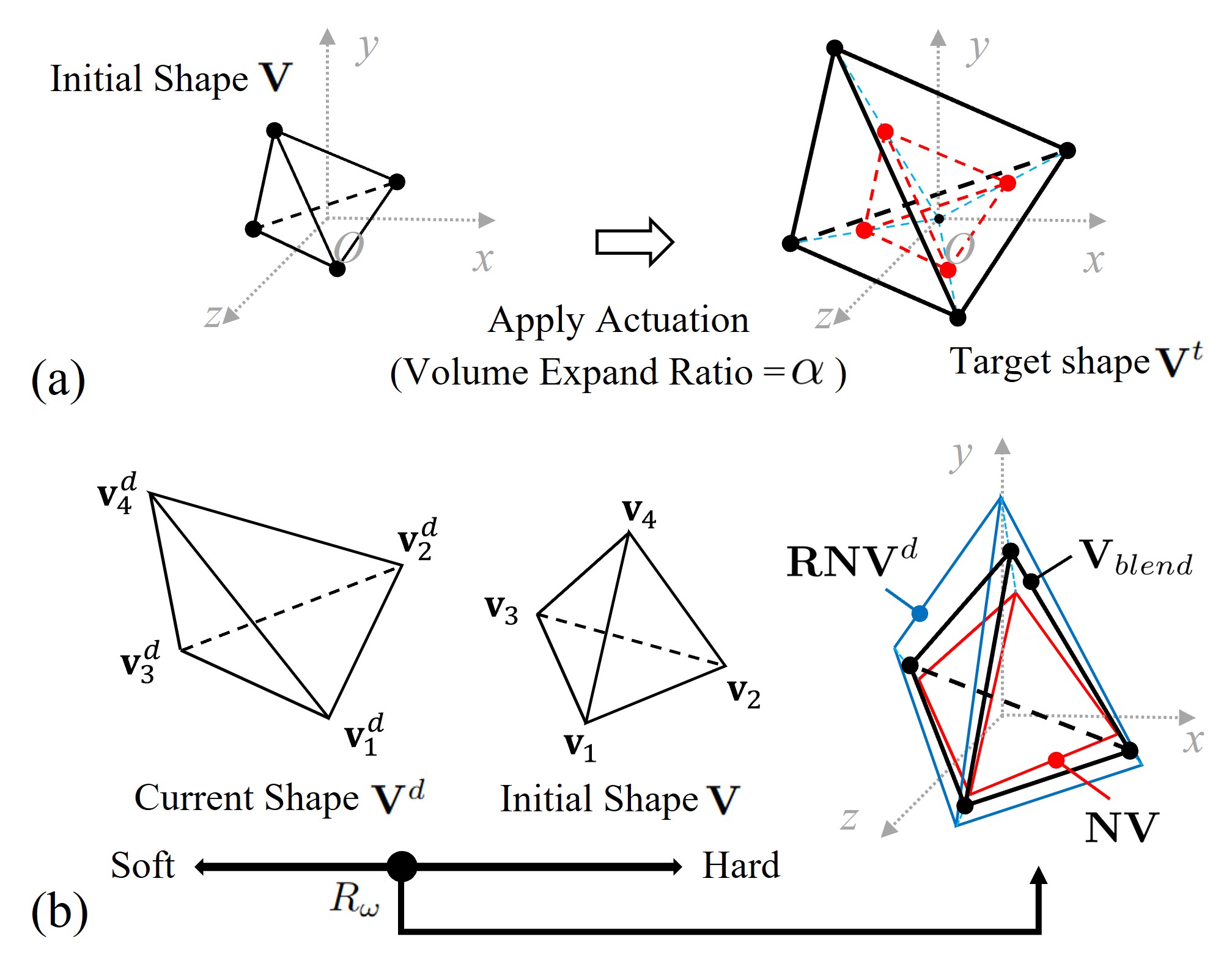}\\
\caption{Computing the target shapes for (a) a chamber element with given actuation parameter $\alpha$ and (b) a body element ($\mathbf{v}\in\mathcal{M}_b$) by incorporating material parameter $R_{\omega}$.}\label{fig:targetShape}
\end{figure}

\subsection{Computing target shapes local projection}\label{subsec:localProjection}
Here we introduce how the target shapes are computed for different types of elements.

\subsubsection{Chamber element with actuation constraint}
The target shapes for chamber elements with volume variation are computed by an average expansion in all verities as illustrated in Fig.~\ref{fig:targetShape}(a). With actuation parameter $\alpha$ as input,  the target position for the $i$-th vertex in the element is determined as $\mathbf{v}_i^t = \sqrt[3]{\alpha} \mathbf{v}_i$. This update ensures that the volume of the chamber region is updated to $\alpha$ times to its initial status that satisfies constraint (Eq.(\ref{eq:actuConst})).

\subsubsection{Body element with hyperelastic material property}
The property of soft material needs to be integrated when computing the target shape for body elements. We modify the scheme of shape blending in~\cite{FANG20_TRO} by inviting strain-related stiffness parameter to mimic the behavior of hyperelasticity in soft material. As illustrated in Fig.~\ref{fig:targetShape}(b), the projected shape is first computed by combing the initial shape and the current shape of an element. 
The target shape is computed by
\begin{equation}\label{eq:blendingElement}
\mathbf{V}_{blend} = (1-R_{\omega}(\varepsilon))\mathbf{RN}\mathbf{V}^d + R_{\omega}(\varepsilon)\mathbf{N}\mathbf{V},
\end{equation}
\begin{equation}\label{eq:volumePreserv}
\mathbf{V}^t = \frac{\mathit{Vol}(\mathbf{V}^t)}{\mathit{Vol}(\mathbf{V}_{blend})} \mathbf{V}_{blend}.
\end{equation}
Here the parameter $R_{\omega}$ is a function of the strain $\varepsilon$ (detail presented in~\cite{Christ21_JCISE}) and can be calibrated from physical test. Meanwhile, to ensure incompressibility of hyperelastic materials (i.e., with Poisson's ratio around 0.5), volume scaling is applied to this element. In our experiments, two silicon rubbers Ecoflex 00-30 and Dargon Skin 20A are modeled since they are widely used in fabricating soft robots~\cite{marchese2015recipe, marchese2016design, yang2020twining}. Detailed data for these materials is provided in Sec.~\ref{sec:result} 
 
\begin{figure}[t] 
\centering
\includegraphics[width=0.9\linewidth]{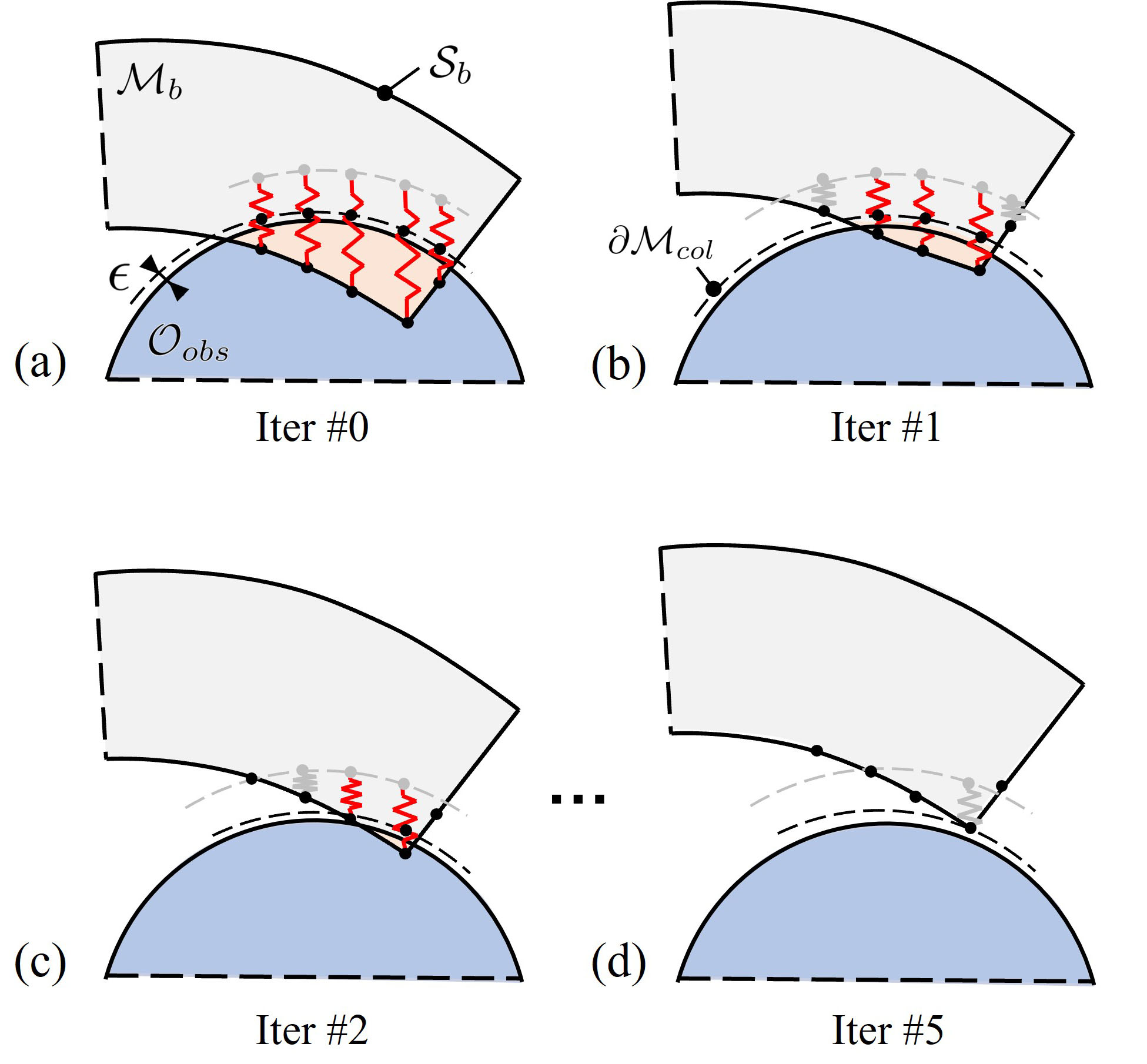}
\caption{Iterative updating spring elements and their correspondences $\mathbf{v}_c$ in collision response. (a) Initial status contains spring elements added on detected collision vertices. (b, c) Through the iteration, the spring elements are deactivated (shows in gray) when $\mathbf{v}$ has been pulled out of collided region $\mathcal{M}_{col}$. (d) Final collision-free status with all spring elements released.
%Iterative updating spring elements and their correspondences $\mathbf{v}_c$ in collision response. The gray color spring present the deactivated spring elements when $\mathbf{v}$ has been pulled out of collided region $\mathcal{M}_{col}$.
}\label{fig:springUpdate}
\end{figure}

\subsubsection{Spring element with updated correspondence}
As explained in Sec.~\ref{subsec:collResponse}, the target shape of a spring element is set as the spring's initial status (i.e., letting the collided point $\mathbf{v}$ move to its corresponding point $\mathbf{v}_c$). In each iteration, the target shape for a spring element needs to be updated together with collision detection. This process is illustrated and explained in Fig.~\ref{fig:springUpdate}. An existing spring element will be released if the vertex $\mathbf{v}$ has been outside the collided region and is away from the boundary of $\mathcal{M}_{col}$. That is
\begin{equation}\label{eq:ColSpringCondition}
\mathbf{v} \notin \mathcal{O}_{obs}, \ 
        ||\vec{\textbf{n}} \cdot (\mathbf{v} - \mathbf{v}_c)|| > \epsilon.
\end{equation}
As an example, springs shown in Fig.~\ref{fig:springUpdate}(b) with a gray color are to be removed in the next iteration since they have satisfied the mentioned condition (Eq.(\ref{eq:ColSpringCondition})). The iteration of collision response terminates when all the springs are released (i.e., $\mathcal{V}_{col} = \emptyset$ and see Fig.~\ref{fig:springUpdate}(d) for an example).

\subsection{Local-global solver with progressive remeshing}
After defining the target shapes for all elements, the optimization problem can be transformed to an augmented form that minimizes the energy
\begin{equation}
    \sum_{e\in \mathcal{M}_b, \mathcal{M}_c} \omega_e ||\mathbf{N}\textbf{V}^d - \mathbf{R} (\mathbf{N}\textbf{V}^t)||^2 + \sum_{\mathbf{v}\in\mathcal{V}_{col}} ||\mathbf{v}-\mathbf{v}_c||^2. \label{eq:finalEnergy}
\end{equation}
By adopting the strategy of local-global solver, the variables $\mathbf{V}^t$, $\mathbf{R}$, and $\mathbf{v}_c$ are computed in the \textit{local step} by shape projection. They will be fixed in the \textit{global step} of blending. Therefore, the energy minimization problem defined in Eq.(\ref{eq:finalEnergy}) becomes a least-square problem as only the position of vertices in $\{ \textbf{V}^d\}$ are unknown variables.
%In this \textit{global step}, the deformed shape $\mathcal{M}^d$ is computed. This in turn requires those shape-related variables to be updated accordingly otherwise the objective energy will increase. 
The local and global steps are applied in an iterative manner, and it has been proved that this solver can effectively minimize the energy in a few steps. In our implementation, the Anderson-Acceleration method~\cite{peng2018anderson} is applied to further improve the converging speed of numerical computations.

\begin{figure}[t] 
\centering
\includegraphics[width=0.9\linewidth]{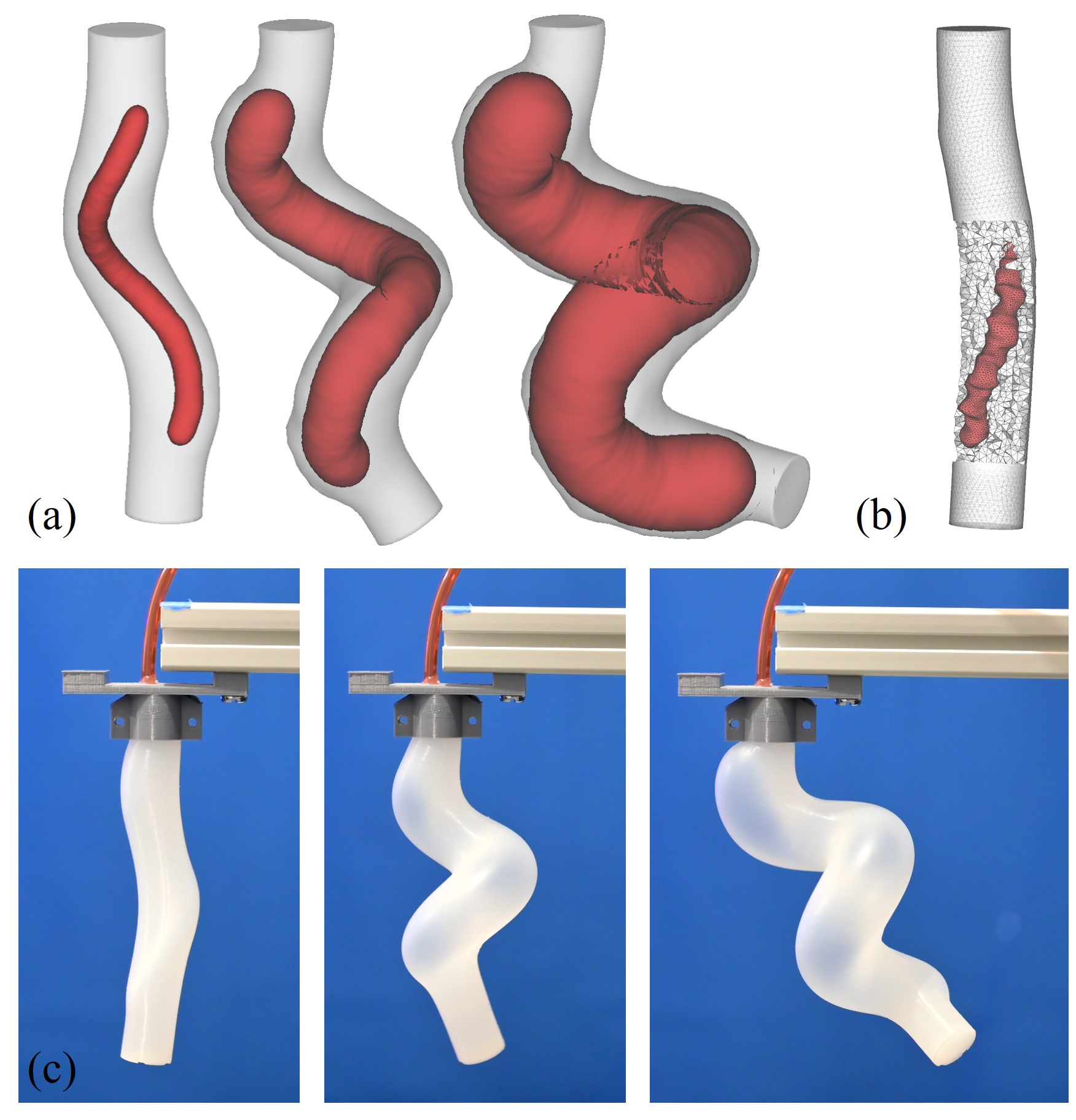}
\caption{Simulation results for a soft robot with twisting deformation under actuation. (a) By progressively remeshing the chamber region $\mathcal{M}_c$ (shows in red), our method can generate the results that mimic the physical behavior shown in (c) as applying different volume-based actuation parameters (from left to right: $\alpha = 5, 28$, and $115$). (b) Directly applying a large actuation parameter $\alpha = 28$ leads to an unrealistic result in simulation.}\label{fig:twistSoftRobot}
\end{figure}

When directly applying a large volume-variation based actuation in the simulation solver, some unrealistic results could be obtained (see Fig.~\ref{fig:twistSoftRobot}(b) for an example). This is caused by a very sparse density of elements in the region of chamber $\mathcal{M}_c$ in the robot's initial shape. To tackle this issue, we progressively apply actuation and remesh the chamber region $\mathcal{M}_c = \mathcal{T}(\mathcal{S}_c^d)$ when necessary. Specifically after deforming a soft robot, we apply the remeshing step if any of the chamber elements is detected either
\begin{enumerate}
\item having $\mathit{Vol}(\mathbf{V}^d) / \mathit{Vol}(\mathbf{V}) > \alpha_{max}$ as a significant volume change, or

\item with $\mathit{norm}(\mathbf{\Sigma}) > d_{max}$ reflecting tremendous distortion in its shape.
\end{enumerate}
Here $\mathbf{\Sigma}$ is the diagonal scaling matrix obtained using the SVD of the affine transformation matrix from $\mathbf{V}^d$ to $\mathbf{V}$. The threshold values $\alpha_{max} = 4.0$ and $d_{max}= 4.0 $ are selected based on the experiments.

An example is presented in Fig.~\ref{fig:twistSoftRobot} to demonstrate the importance and effectiveness of this remeshing step, where a pneumatic-driven twisting soft robot can have the chamber region expanded more than 100 times in comparison to its initial volume. By applying progressive remeshing, the simulation result of a twisting soft robot can realistically match with the result of physical experiments (illustrated in Fig.~\ref{fig:twistSoftRobot}(c)). The local-global solver is illustrated in \textbf{Algorithm 1}, in which the remeshing and collision response are presented.
%In the next section, we present the simulation result of different soft robot design by applying the proposed method. Statics analysis on the computing speed is implicitly presented.

\begin{algorithm}[t]\label{algSolver} 

\caption{Collision-Aware Soft Robot Simulation}

\LinesNumbered

\KwIn{Soft robot model $\mathcal{S}_c$ and $\mathcal{S}_b$, actuation parameter $\alpha$, obstacle model $\mathcal{O}_{obs}$.}

\KwOut{Deformed collision-free soft robot body shape $\mathcal{S}_b$.} 

Generate FE model $\mathcal{M} = \mathcal{M}_c + \mathcal{M}_b$ by $\mathcal{T}(\mathcal{S}_c,\mathcal{S}_b)$;

$k = 0$, $\mathcal{\alpha}_k = \alpha_{max}, \mathcal{M}_c^k = \mathcal{M}_c$; 

\While{$\mathit{Vol}(\mathcal{M}_c^k) < \alpha\mathit{Vol}(\mathcal{M}_c)$}
{

\tcc{\footnotesize Collision-aware local-global solver}

\While{$\mathcal{V}_{coll} \neq \emptyset$ and $i<i_{max}$}{

    $\mathbf{V}^t, \mathbf{R}, \mathbf{v}_c \leftarrow \textit{local-step} (\mathcal{V}_{coll}, \mathcal{M}^{i}, \alpha_k)$;
 
    $\mathcal{M}^{i+1} \leftarrow \textit{global-step}(\mathbf{V}^t, \mathbf{R}, \mathbf{v}_c$);
    
    update $\mathcal{V}_{coll} \leftarrow \textit{CollisionCheck}(\mathcal{M}^i, \mathcal{O}_{obs})$;

}

\tcc{\footnotesize Check if remeshing needed
}

\For{$e \in \mathcal{M}_c$}{

\If{\footnotesize $\mathit{Vol}(e_{k+1}) / \mathit{Vol}(e_{k}) > \alpha_{max}$ or $\mathbf{\Sigma}(e) > d_{max}$ }{
Remesh and update $\mathcal{M}^k = \mathcal{M}_b^k + \mathcal{T}(\mathcal{S}_c^k)$;
}

}

$k = k+1$, and $\alpha_k = k\alpha_{max}$;

}

\textbf{return} $\mathcal{S}_b = \partial \mathcal{M}_b^k$;

\end{algorithm}

%Notice that this local-global solver (line 5-7) is proved to reduce the energy function value in each iteration and can be accelerated by Anderson Acceleration where a proper time step can be find for each step can be optimized. This can improve the converge speed of the algorithm that accelerate the simulation speed of the system. The AA-based acceleration is also inflected in the algorithm and the comparison in its performance is present in the result section below.

%\input{tex/secLearning}
%\input{tex/secPlanning}
\section{Case Studies and Experiment Results} \label{sec:result}
In this section, we demonstrate the behavior of our collision-aware soft robot simulator by presenting several case studies. The proposed computational framework was implemented with C++ and run with a PC with AMD 3950x CPU and 32GB memory. The parallel computing was supported by OpenMP, and the Eigen library was used to solve the linear system. The simulation results were also compared to physical experiments to verify the performance of the approach.

\subsection{Twisting soft robot}

% \begin{figure}[t]
% \centering
% \includegraphics[width=1.0\linewidth]{figure/twistingResult.jpg}
% \caption{Simulation result for soft robot with twisting behavior. \guoxin{To replace the x-axis as the pressure inside the chamber. Also add the physical experiment figure for comparison.}}\label{fig:twistingShape}
% \end{figure}
\begin{figure}[t]
\centering
\includegraphics[width=1.0\linewidth]{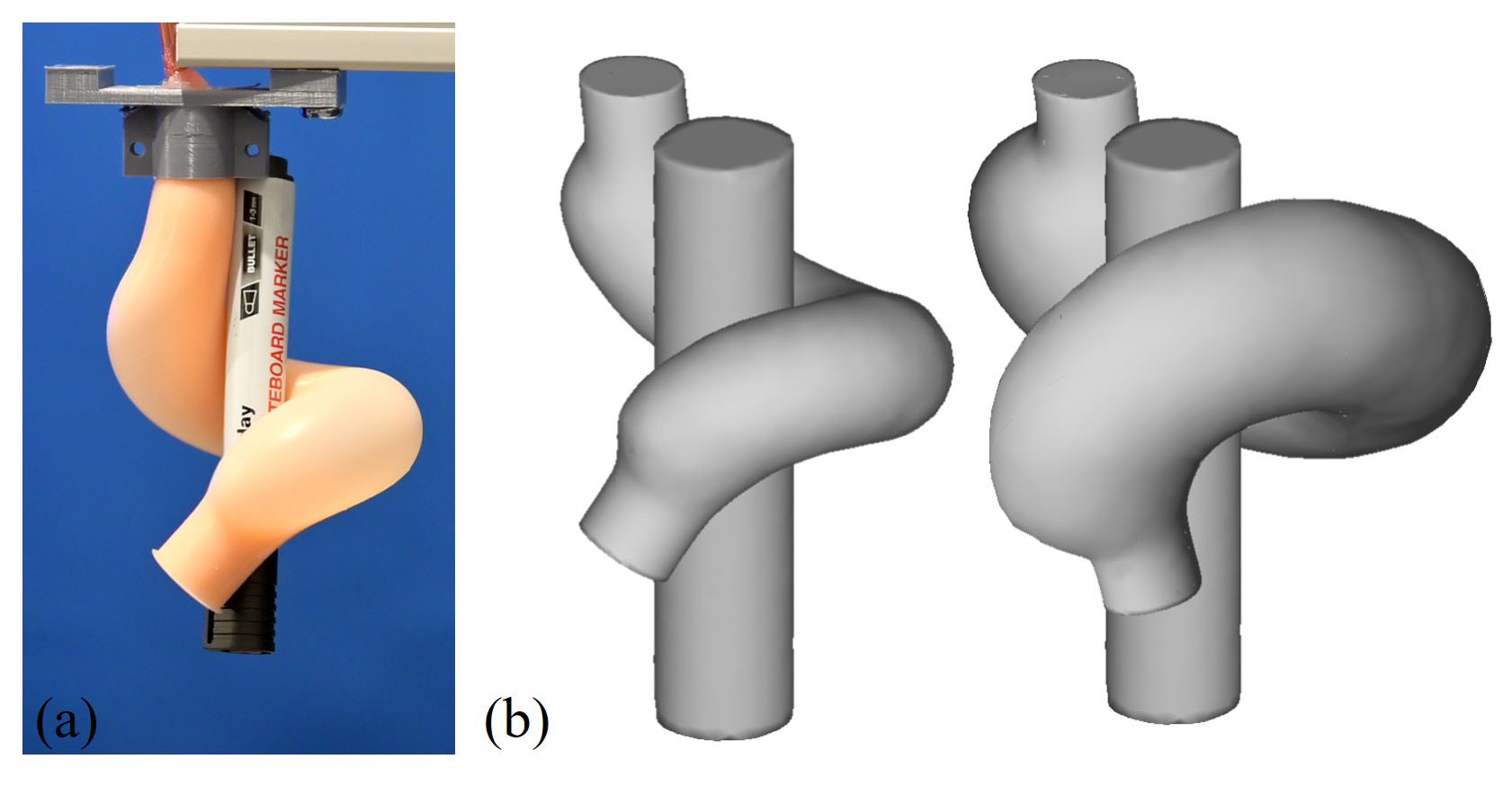}
\caption{Case study taken on a twisting soft robot -- (a) physical experiment of using the twisting soft robot to hold a cylindrical object, and (b) our simulator can successfully model the collision behavior when the obstacle is added and when large pneumatic actuation is applied.}\label{fig:twistInteract}
\end{figure}

The first case study is conducted on a plant inspired soft robot~\cite{yang2020twining} that can perform twisting deformation. The robot consists of a cylindrical body and a spiral-shape chamber. Eco-Flex 00-30 silicon rubber is used to fabricate the robot, and we model its hyperelastic material property by using first-order Mooney-Rivlin with $c_{1}$ = 0.0418 MPa and $c_{2}$ = 0.0106 MPa~\cite{yang2020twining}. The simulation results are presented in Fig.~\ref{fig:twistSoftRobot}(b) and can mimic the physical behavior well when remeshing is applied to the chamber region. 
%
%Compared with other method, our method is fast and performs the best in convergence (see comparsion in Table I). The full FEA model tested on Abaqus software and reduced SOFA simulator failed to converge when large deformation happen. 
%

This twisting soft robot is able to firmly twine a target object. As depicted in Fig.~\ref{fig:twistInteract}, a cylindrical obstacle is tightly grasped when the soft robot is actuated into a shape around the cylinder. This behavior is also successfully simulated by our method, where larger actuation can be further added to the chamber region to produce a tighter grasping action (see the right of Fig.~\ref{fig:twistInteract}(b)). The mesh model used for simulating this soft robot contains 28.2k tetrahedra. With this size of mesh model, our simulator can achieve a computing speed of 8.09 seconds per time step. This is around 25 times faster than a FEA-based simulation (conducted by Abaqus software) that takes more than 3 minutes on a mesh with a similar density. Commercial software such as Abaqus also suffers from the convergence problem due to the distortion in elements when large inflation is applied (details can be found in Case II in the supplementary video).
%\subsection{Multi-chambers bending actuator}
%\guoxin{The design mainly from IJRR 2015, present the result of actuation different chambers 1 (10\%, 30\%, 50\%, 100\%), 2, and 4. Also add the comparison of including hyper-elastic model inside the simulation.}

\begin{figure}[t]\centering
\includegraphics[width=1.0\linewidth]{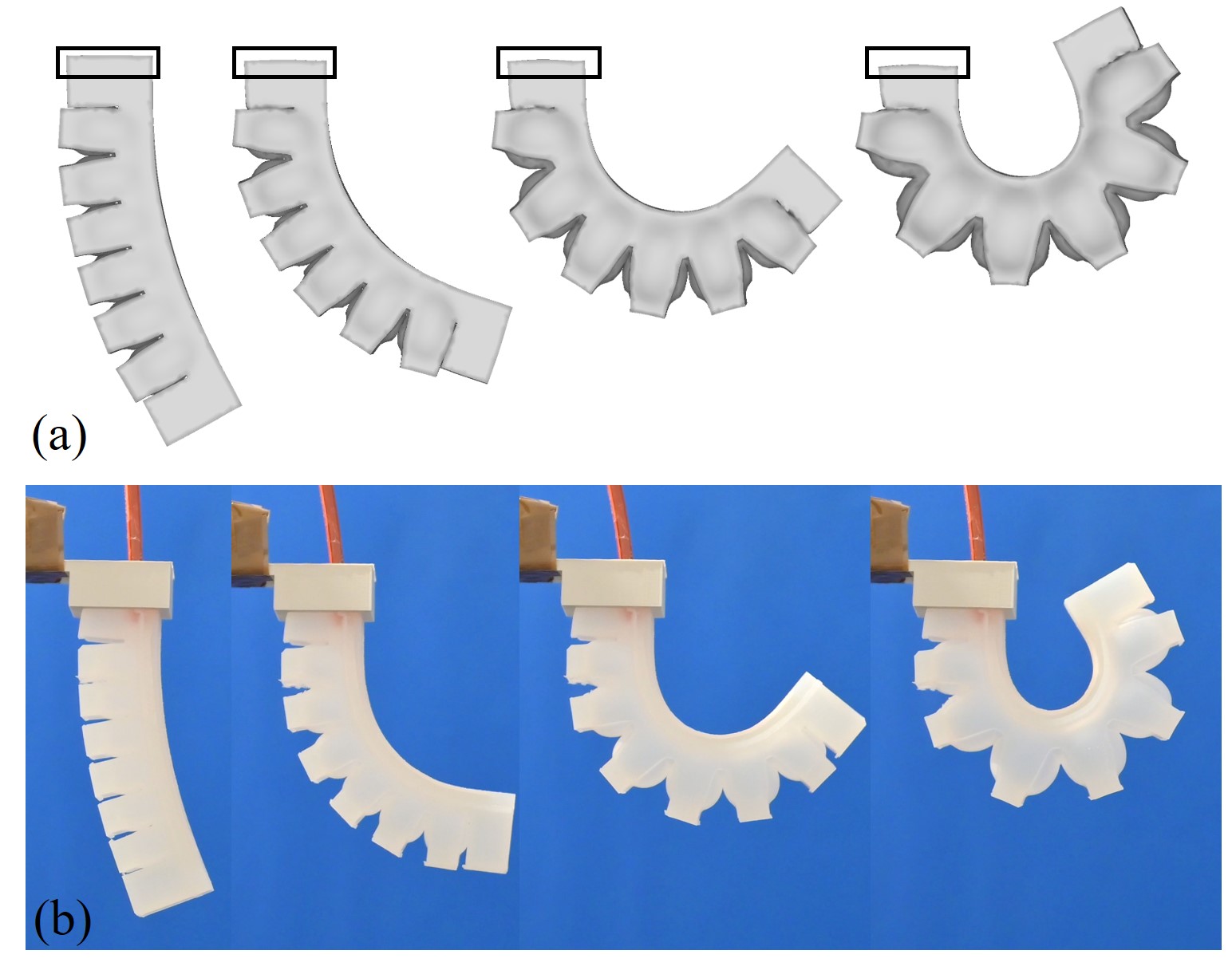}
\caption{(a) Simulation result for the soft finger made by silicone rubber and can perform large bending deformation. (b) The simulation result can well capture the physical behavior.}\label{fig:fingerBending}
\end{figure}

\subsection{Soft finger with self-collision}
\begin{figure}[t]\centering
\includegraphics[width=0.9\linewidth]{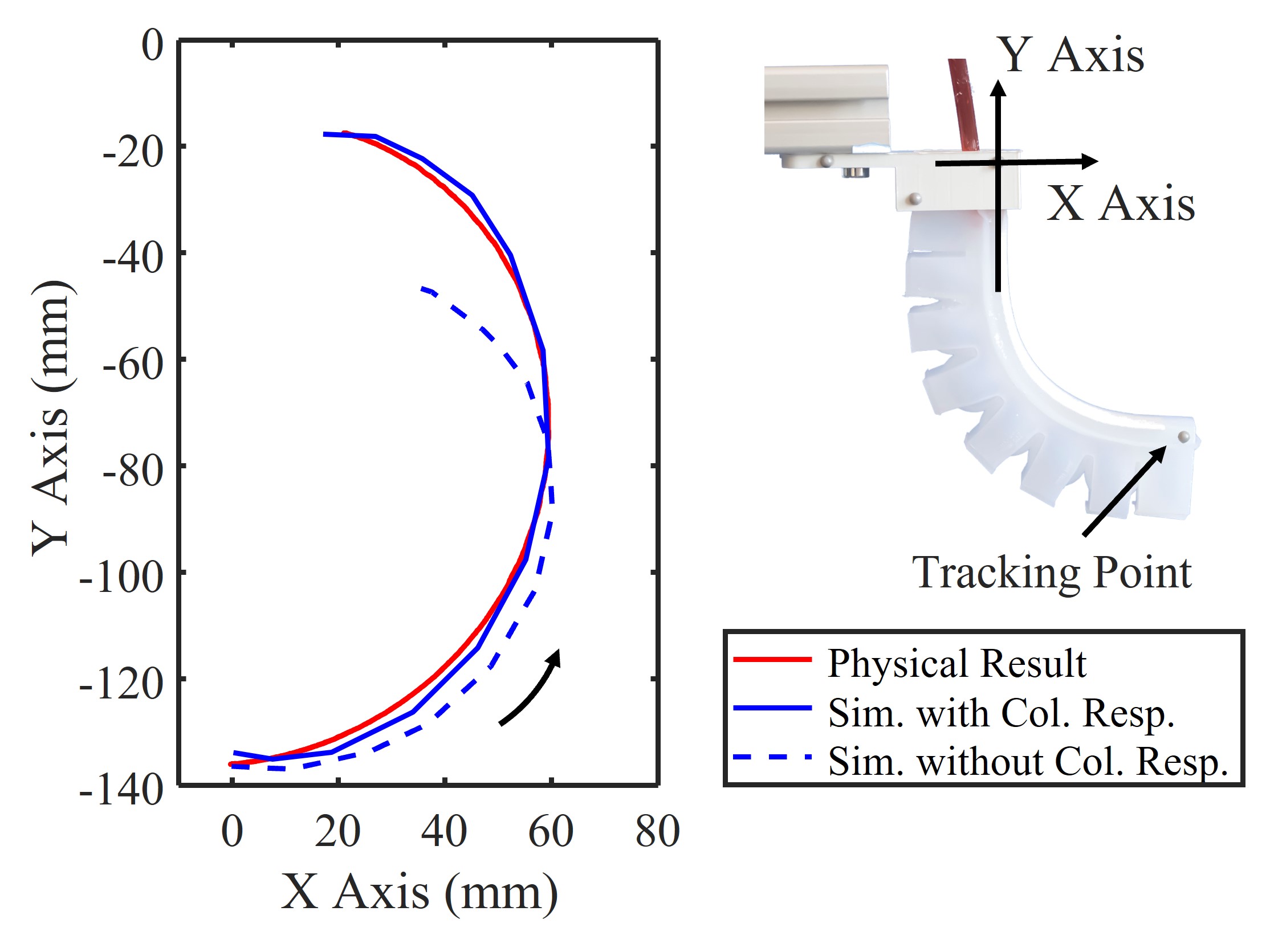}
\caption{The simulator can predict the tip trajectory of the soft finger more precisely after integrating the collision response model.}\label{fig:fingerTraj}
\end{figure}

The second case study is conducted on a soft finger model, where both self-collision and robot-obstacle interactions occur at the same time. The result of free-bending considering self-collision between chambers is presented in Fig.~\ref{fig:fingerBending}, and the comparison on the tracked tip position can be found in Fig.~\ref{fig:fingerTraj}. Without an effective collision response, the simulator presented in our previous work~\cite{FANG20_TRO}
%and SOFA~\cite{duriez2013control} 
cannot precisely mimic the behavior of soft finger (i.e., large tracking error occurs in the tip trajectory). By using the collision-aware method presented in this paper, the error is reduced.

\begin{figure}[t]\centering
\includegraphics[width=1.0\linewidth]{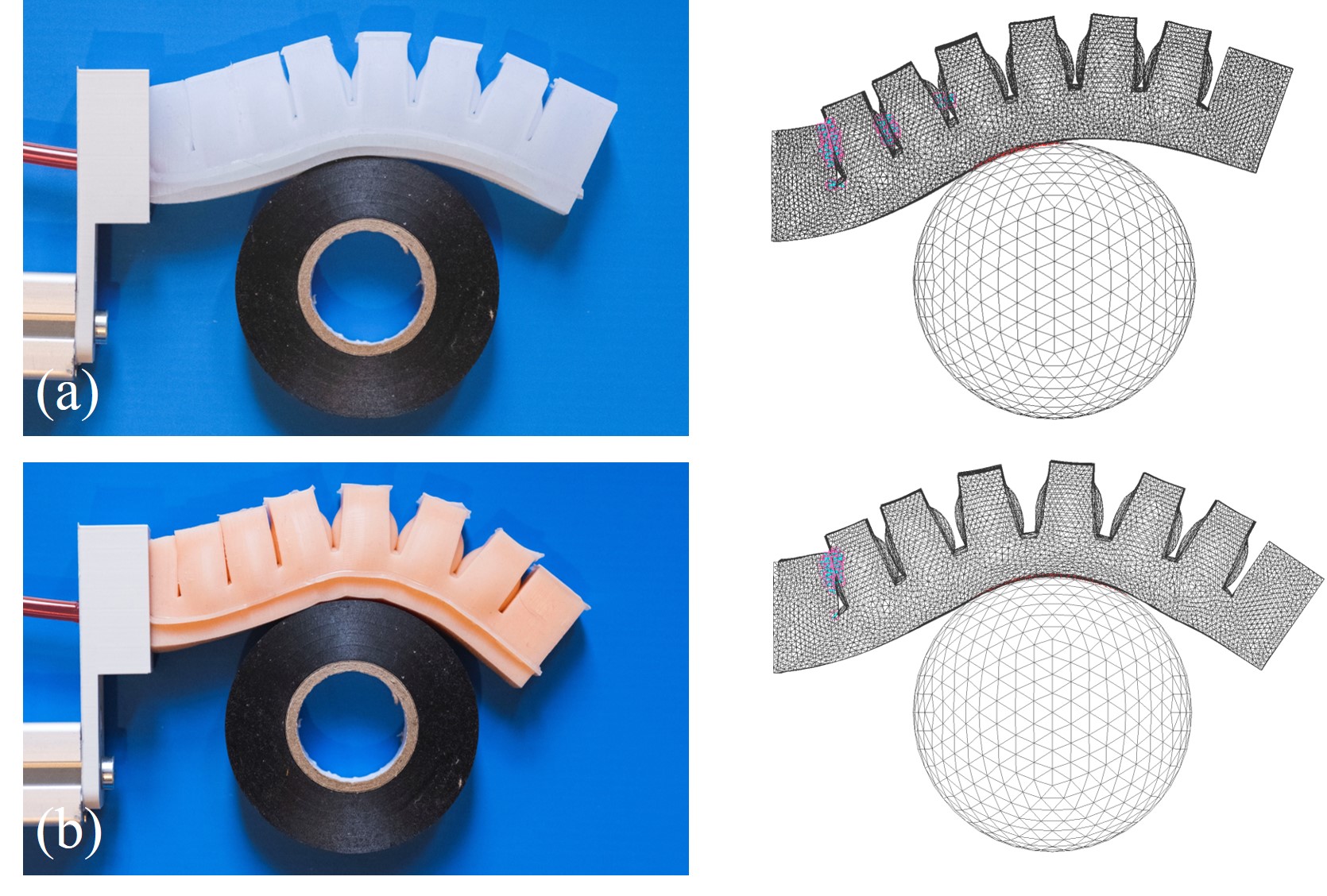}
\caption{Behavior of soft grippers while interacting with an obstacles with round shape. The soft grippers were fabricated using different types of silicon rubber, such as (a) Dragon Skin 20 and (b) Eco-flex 00-30. Our simulator results (depicted on the right) successfully predict both behaviors with corresponding material properties, respectively. }\label{fig:materialCompare}
\end{figure}

We also use two soft materials, Eco-flex 0030 and Dargon Skin 20, to fabricate two soft fingers for comparison. The Dargon Skin 20 material is around 3 times stiffer, and the Mooney–Rivlin model with $c_1 = 0.119$ MPa, $c_2 = 0.023$ MPa is used. As shown in Fig.~\ref{fig:materialCompare}, two soft robots perform differently when interacting with obstacles. With a softer material, the contact region becomes larger and the resultant shape is more around the obstacle's boundary. This phenomenon of different materials is well captured by our simulator with the incorporated hyperelastic material model.

\begin{figure}[t]
\centering
\includegraphics[width=1.0\linewidth]{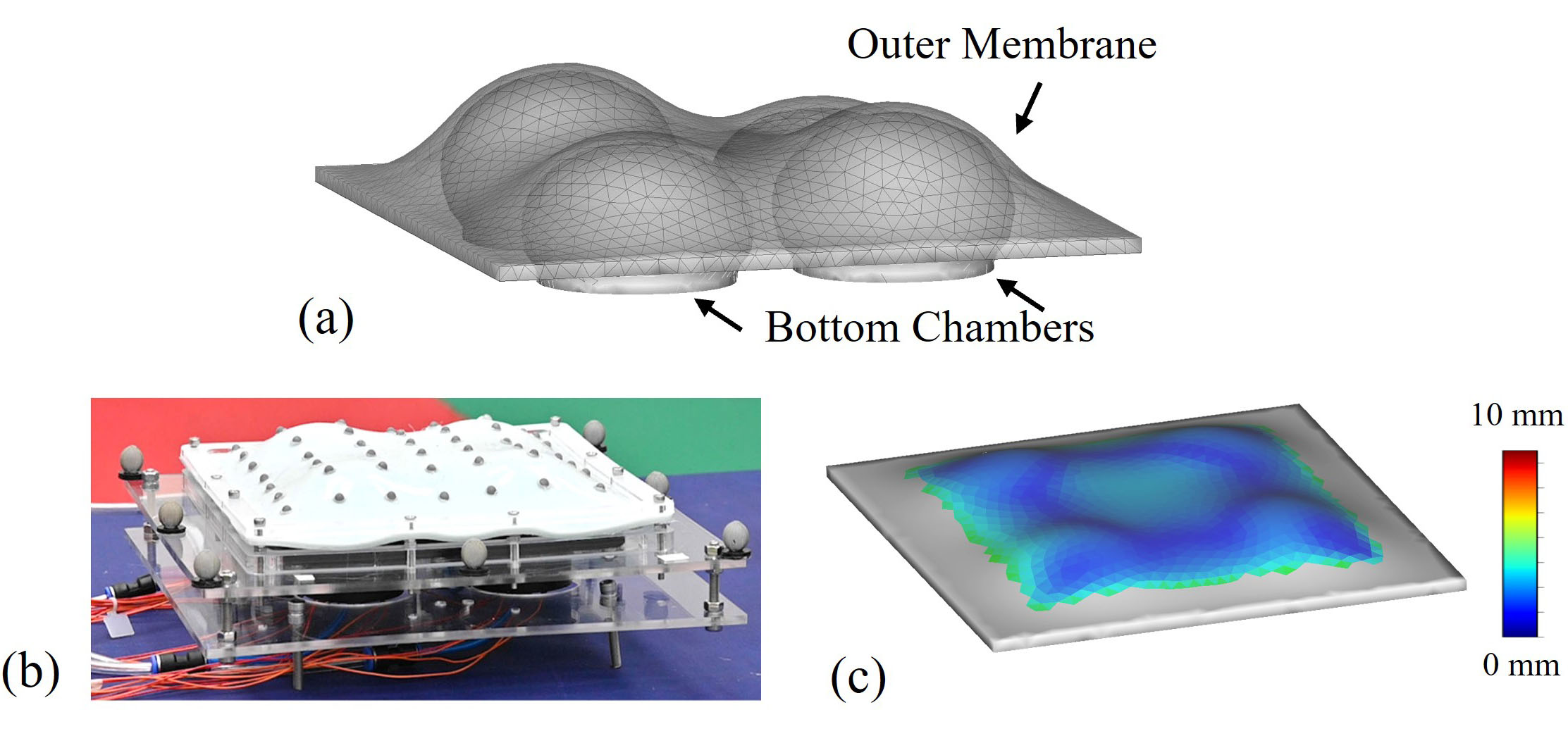}
\caption{Soft membrane driven by interaction with chambers at the bottom -- (a) simulation result, (b) physical result on hardware, and (c) quantities analysis of the shape estimation error.}\label{fig:mannequin}
\end{figure} 

\vspace{-5px}

\subsection{Soft membrane driven by interactions}
To verify the behavior in interactions between multiple soft robots, the proposed method is applied to simulate the behavior of a soft membrane setup as illustrated in Fig.~\ref{fig:mannequin}. The membrane can be deformed to a free-form shape through interactions with four pneumatic-driven chambers below it~\cite{scharff2021sensing, Tian22TMECH}. The size of the setup is $250 \times 250$ mm, which is fabricated using Dragon Skin 20. Our model precisely captures the shape of the outer membrane according to the given actuation parameters applied to the inner chamber array. The predicted shape is compared to the shape captured from physical setup by motion capture system. The maximum error is 5.0 mm as shown in Fig.~\ref{fig:mannequin}, which is 2\% of the hardware setup's size. The simulated motion of the membrane can also be found as case IV in the supplementary video.

\subsection{Discussion and Limitation}

\begin{table}[]
\centering
\caption{Computational cost of the collision-aware simulator$^\dag$}
\begin{tabular}{c|ccccc}
\hline \hline
\specialrule{0em}{2pt}{1pt}
Models & Tet. \# & \begin{tabular}[c]{@{}c@{}}Mesh \\ Gener.\end{tabular} & \begin{tabular}[c]{@{}c@{}}Col. \\ Check \end{tabular} & \begin{tabular}[c]{@{}c@{}}Col. \\ Resp.\end{tabular} & \begin{tabular}[c]{@{}c@{}}Iter. \\ Solver\end{tabular} \\ \specialrule{0em}{1pt}{1pt} \hline \specialrule{0em}{1pt}{1pt}
Soft Finger & 79,048 & 3.54 s & 0.07 s & 4.38 s & 7.46 s \\ 
\specialrule{0em}{1pt}{1pt}
Twisting Soft Robot & 28,198 & 2.81 s & 0.03 s & 1.71 s & 3.54 s \\  \specialrule{0em}{1pt}{1pt}
Soft Membrane & \ 34,680$^*$  & 0.96 s & 0.12 s & 1.63 s & 3.83 s\\ \specialrule{0em}{1pt}{1pt} \hline \hline
\end{tabular}
\begin{flushleft}
$^\dag$The computing time is for single step with given actuation input. \\
$^*$Contains four chamber, each one with 8,670 tetrahedrons.
\end{flushleft}
\label{tab:compTime}
\end{table}

The statistics of computing costs are listed in Table~\ref{tab:compTime}. In general, the computing speed of our simulator is around 25 times faster than the FEA-based simulation on the mesh models with similar density. In comparison to other reduced FEA-based numerical models (e.g., SOFA~\cite{duriez2013control}) that have faster computing speeds, our method is more robust when large nonlinear deformations happen, which guarantees converging with a realistic simulation result. In all steps of our simulation, solving the optimization problem Eq.(\ref{eq:finalEnergy}) takes more than 45\% of the simulation time. Collision detection is run in real time (i.e., less than 0.12 seconds for all cases), and collision response consumes 25\% of the total time. The time used to generate tetrahedral mesh can be further reduced by inviting a more efficient method for volume tessellation~\cite{labelle2007isosurface}.
 
The major limitation of our current work is that the prediction of contact force is not included in the proposed collision-response model. One solution could be first computing the strain for each element based on the computed deformation, and then modelling the external force by strain-stress relationship with force equivalence function. 
This can be an interesting future work. Meanwhile, only the friction-less contact is considered in this work -- see the soft robot sliding on the contact plan as demonstrated in Case I of the supplementary video. It's in general hard to model frictions precisely since the stick/slip condition brings a non-smooth collision response process that is hard to be computed efficiently.

\section{Conclusion}\label{sec:conclusion}
In this paper, we develop a collision-aware soft robot simulator based on geometric computing. Compared with existing FEA-based approaches, the proposed method demonstrates a much faster computational speed and very robust convergence with complex deformation in a robot's body shape (e.g., expanding, twisting, and bending). Both self-collision and robot-obstacle interactions are successfully modeled. We have tested the simulator on a variety of soft robot systems fabricated from different materials with hyperelastic property, where our results can accurately simulate the physical deformation of soft robots.

% \clearpage
{\small
\bibliographystyle{IEEEtran}
\bibliography{references}
}

\end{document}